\documentclass[10pt,twocolumn,letterpaper]{article}

\usepackage[pagenumbers]{cvpr} 

\usepackage{graphicx}
\usepackage{amsmath}
\usepackage{amssymb}
\usepackage{booktabs}
\usepackage{multirow}
\usepackage{tabularx}
\usepackage{color}
\usepackage{enumerate}
\usepackage{enumitem}
\usepackage[misc]{ifsym}
\usepackage{bbding}
\usepackage{bbm}

\usepackage[pagebackref,breaklinks,colorlinks]{hyperref}

\usepackage[capitalize]{cleveref}
\crefname{section}{Sec.}{Secs.}
\Crefname{section}{Section}{Sections}
\Crefname{table}{Table}{Tables}
\crefname{table}{Tab.}{Tabs.}

\def\cvprPaperID{3303} 
\def\confName{ICCV}
\def\confYear{2023}

\begin{document}

\title{Efficient Teacher: Semi-Supervised Object Detection for YOLOv5}

\author{Bowen Xu \quad Mingtao Chen \quad Wenlong Guan \quad Lulu Hu\\
Alibaba Group\\
{\tt\small \{bowen.xbw,ruiyang.cmt,wenlong.gwl,chudu.hll\}@alibaba-inc.com} \\
}

\maketitle

\begin{abstract}
Semi-Supervised Object Detection (SSOD) has been successful in improving the performance of both R-CNN series and anchor-free detectors. However, one-stage anchor-based detectors lack the structure to generate high-quality or flexible pseudo labels, leading to serious inconsistency problems in SSOD. In this paper, we propose the Efficient Teacher framework for scalable and effective one-stage anchor-based SSOD training, consisting of Dense Detector, Pseudo Label Assigner, and Epoch Adaptor.
Dense Detector is a baseline model that extends RetinaNet with dense sampling techniques inspired by YOLOv5. The Efficient Teacher framework introduces a novel pseudo label assignment mechanism, named Pseudo Label Assigner, which makes more refined use of pseudo labels from Dense Detector. Epoch Adaptor is a method that enables a stable and efficient end-to-end SSOD training schedule for Dense Detector. The Pseudo Label Assigner prevents the occurrence of bias caused by a large number of low-quality pseudo labels that may interfere with the Dense Detector during the student-teacher mutual learning mechanism, and the Epoch Adaptor utilizes domain and distribution adaptation to allow Dense Detector to learn globally distributed consistent features, making the training independent of the proportion of labeled data.
Our experiments show that the Efficient Teacher framework achieves state-of-the-art results on VOC, COCO-standard, and COCO-additional using fewer FLOPs than previous methods. To the best of our knowledge, this is the first attempt to apply SSOD to YOLOv5.
\end{abstract}

\section{Introduction}
\label{sec:intro}

Object detection\cite{lin2017focal,ren2015faster,tian2019fcos,cai2018cascade} has made significant advances in recent years, which follows a traditional supervised training approach and relies on costly manual annotation efforts. To mitigate this problem, many semi-supervised techniques\cite{sohn2020fixmatch,berthelot2019mixmatch} are proposed to exploit large amounts of unlabeled data by automatically generating pseudo labels without introducing manual annotation. Despite great progress in SSOD\cite{chen2022label,chen2022dense,xu2021end,liu2021unbiased}, there are three key issues that remain challenging:
\begin{figure*}[ht]
    \centering 
    \includegraphics[width=160mm,height=85mm]{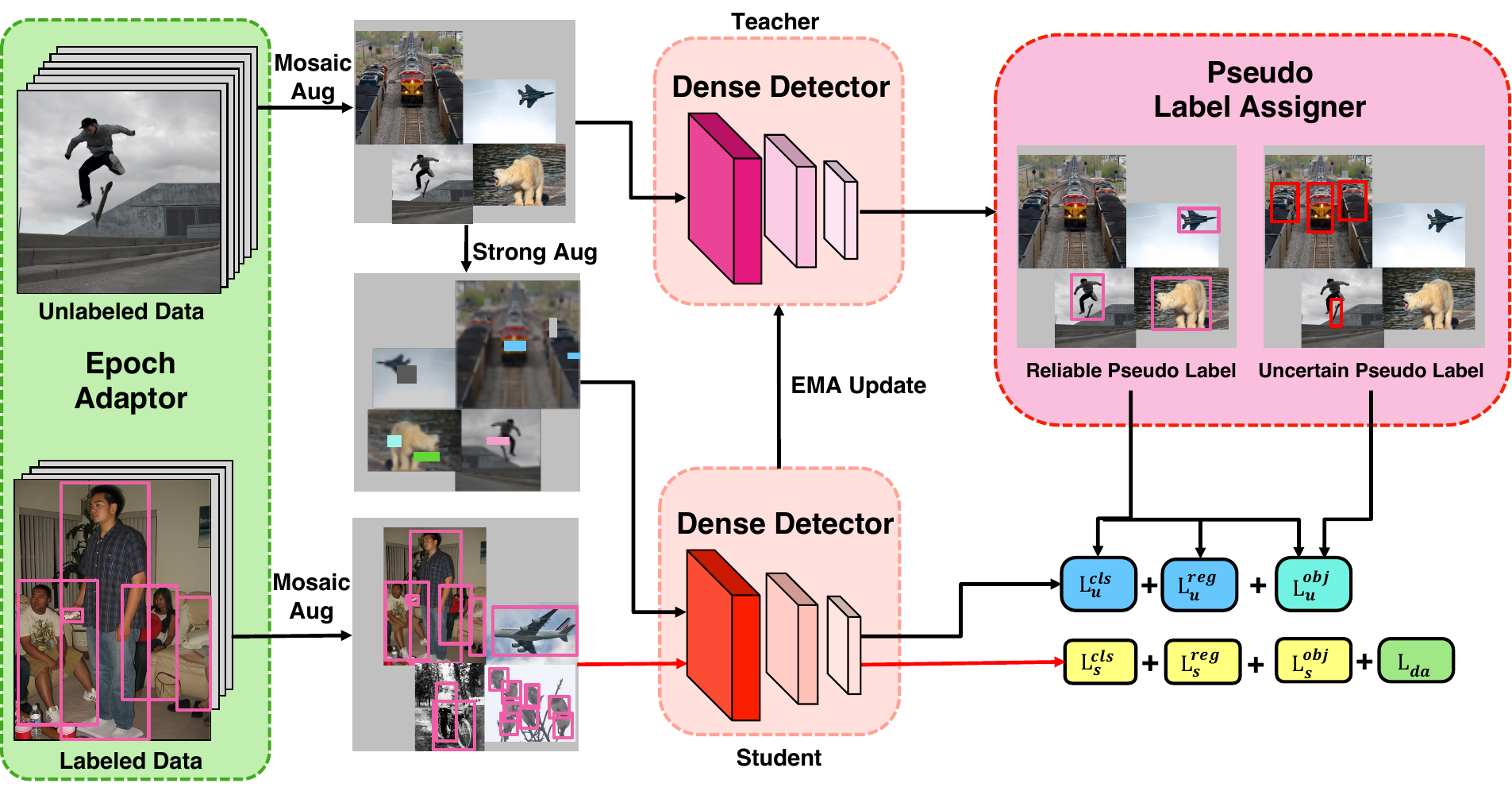}
    \caption{An overview of Efficient Teacher framework. Efficient Teacher proposes three modules to implement a scalable and
    effective SSOD framework, where Dense Detector improves the quality of pseudo labels with dense input while has better inference efficiency; Pseudo Label Assigner divides pseudo labels into two types to alleviate pseudo labels inconsistency problem; Epoch Adaptor reduces training time and the inconsistency of features.}
    \label{Fig.efficient_teacher_arch}
\end{figure*}

Firstly, \textbf{few works on one-stage anchor-based SSOD have been reported}. Though anchor-free detectors\cite{ge2021yolox, tian2019fcos, li2022yolov6}have been recently getting
more attention in the community of object detection, one-stage anchor-based detectors \cite{redmon2018yolov3,bochkovskiy2020yolov4,jocher2022ultralytics,li2022yolov6,wang2022yolov7}, having the advantages of high recall, high numerical stability and fast training speed,  are widely used in scenarios with extremely high recall demands. However, most SSOD methods are implemented on a two-stage anchor-based detector such as Faster R-CNN\cite{ren2015faster} and an one-stage anchor-free detector such as FCOS\cite{tian2019fcos}, which output relatively sparse bounding box predictions due to the multi-stage coarse-to-fine prediction mechanism or the anchor-free design of detection head. In contrast, the classic one-stage anchor-based detector generates more dense predictions due to 
its multiple-anchor mechanism, which leads to positive and negative samples imbalance during supervised training\cite{zhang2020bridging} and poor quality of pseudo labels during semi-supervised training.

Secondly, current mainstream SSOD approaches, following a teacher-student mutual learning manner\cite{liu2021unbiased,xu2021end}, is difficult for an one-stage anchor-based detector to train due to the serious \textbf{pseudo label inconsistency} problem, that is, throughout the training process, the quantity and quality of pseudo labels generated by the teacher model fluctuates greatly and the unqualified pseudo labels can mislead model updates.
To alleviate this problem, two-stage methods\cite{chen2022label}\cite{liu2021unbiased} refine pseudo labels several times more than one-stage methods 
and anchor-free methods\cite{zhou2022dense} adopt feature maps as soft pseudo labels to avoid bias caused by non maximum suppression.
The pseudo label inconsistency is exacerbated in an one-stage anchor-based detector because of its multiple-anchor mechanism mentioned above.  
The work\cite{zhang2022s4od} has reported that the SSOD experimental results of RetinaNet are not as good as those on Faster R-CNN and FCOS.

Thirdly,\textbf{ how to train a SSOD model with both higher accuracy and better efficiency} becomes the key issue that restricts the application of SSOD in a wide range of scenarios. The previous SSOD methods\cite{xu2021end, liu2021unbiased,chen2022label,zhou2021instant,li2022pseco} are mainly in pursuit of better accuracy, but usually sacrifice training efficiency. Moreover, most previous works only focus on specific detector architecture, but the variety of real-world applications require faster iterative detector design with lower compute resource and higher accuracy. 

In this paper, what we pursue is to design a scalable and effective SSOD framework on an one-stage anchor-based detector while considering both inference and training efficiency. We add the effective techniques used in the YOLO series\cite{ge2021yolox, bochkovskiy2020yolov4, wang2022yolov7} to a classical RetinaNet\cite{lin2017focal} to design a new representative one-stage anchor-based detector baseline, called Dense Detector. We attempt to transplant a mature SSOD scheme, the Unbiased Teacher\cite{liu2021unbiased}, to Dense Detector but find only 1.65 $AP_{50:95}$ improvement compared to the supervised method(shown in Table~\ref{tab:pseudo label assigner}), which confirms the second problem mentioned above. According to design paradigm of the Dense Detector, we propose the Efficient Teacher framework to overcome these challenges in SSOD. Pseudo Label Assigner(PLA) is introduced to alleviate pseudo label inconsistency by exploiting a fine-grained pseudo label assignment strategy on the objectness branch design. 
By distinguishing the pseudo labels into the reliable and the uncertain regions, different loss calculation methods are used respectively. 
In addition, we propose Epoch Adaptor(EA),  which utilizes domain adaptation and distribution adaptation separately to optimize the training process of the Burn-In phase and SSOD Training phase, respectively. 
Specifically, during the Burn-In phase, EA utilizes domain adaptation techniques for adversarial learning on the output feature maps of the student model. In the SSOD training phase, EA dynamically estimates the threshold for pseudo-labels by online statistics of the proportions of each class label appearing in the labeled data, in order to optimize the quality and distribution of pseudo-labels seen by the student model.The main contributions of this paper are as follows:
\begin{itemize}[leftmargin=12pt, topsep=2pt, itemsep=0pt]
\item We design Dense Detector as a baseline model to compare the differences between YOLOv5 and RetinaNet, which leads to a performance improvement of 5.36 $AP_{50:95}$ by utilizing dense sampling. 

\item We propose an effective SSOD training framework called Efficient Teacher, which includes a novel pseudo label assignment mechanism, Pseudo Label Assigner, reducing the inconsistency of pseudo labels, and Epoch Adaptor, enabling a fast and efficient end-to-end SSOD training schedule. 

\item our experiments demonstrate that utilizing Efficient Teacher on YOLOv5 produces state-of-the-art results on VOC, COCO-standard, and COCO-additional datasets while consuming significantly fewer FLOPs than previous approaches.
\end{itemize}

\section{Related Work}
\label{sec:related work}


\textbf{Semi-supervised Object Detection.} Semi-supervised object detection, inherited from the semi-supervised image classification methods\cite{xie2020self, sohn2020fixmatch, berthelot2019mixmatch, sajjadi2016regularization, tarvainen2017mean}, is divided into consistency-based schemes\cite{jeong2019consistency,tang2021proposal} and pseudo-labeling schemes\cite{liu2021unbiased,sohn2020simple, xu2021end,zhou2021instant}. The latter has become the current mainstream approach. STAC\cite{sohn2020simple} exploits weak and strong data augmentation to process unlabeled data respectively. Unbiased Teacher\cite{liu2021unbiased} follows a stduent-teacher mutal learning to generate more accurate pseudo labels.To balance the effect of pseudo labels, Soft Teacher\cite{xu2021end} uses the scores of the pseudo labels as the weights for loss calculation. DSL\cite{chen2022dense} is the first attempt to perform semi-supervised training on an anchor-free detector(FCOS)\cite{tian2019fcos}. To relieve inconsistency problems, LabelMatch\cite{chen2022label} utilizes label distribution to dynamically determine the filtering threshold of different categories of pseudo labels. The methods above have been proven great performance on two-stage and anchor-free detectors, but can not perform well on an one-stage anchor-based detector. Our Efficient Teacher is proposed to bridge the gap between semi-supervised training and one-stage anchor-based detectors.

\textbf{Label Assignment.} Label assignment is the key component that determines the performance of an object detector. Many works have been proposed to improve the label assignment mechanism, such as ATSS\cite{zhang2020bridging}, PAA\cite{kim2020probabilistic}, AutoAssign\cite{zhu2020autoassign} and OTA\cite{ge2021ota}. 
Some researches\cite{chen2022label}\cite{liu2022unbiased} have noticed that the default label assignment mechanism using in supervised object detection can not be applied in SSOD directly, which results in performance degradation. In this paper, we propose a novel pseudo label assignment that can adapt to SSOD training for one-stage anchor-based detectors.

\textbf{Domain Adaptation in Object Detection.} The task of domain-adaptive object detection\cite{deng2021unbiased,vs2021mega,chen2021self,li2021free}, aims to address the problem of domain shift\cite{chen2018domain}.The work\cite{ganin2015unsupervised} utilizes adversarial learning by training a discriminator with a gradient reverse layer(GRL) to generate domain-invariant feature. The work\cite{deng2021unbiased} introduces semi-supervised techniques used in Mean Teacher to alleviate domain bias, which reveals that domain shift is intrinsically related to inconsistency of semi-supervised task.
This inspires Efficient Teacher to introduce adversarial learning in domain adaptation to alleviate the pseudo label inconsistency of SSOD training.

\section{Efficient Teacher}
\label{sec:Method}
Efficient Teacher is a novel and efficient framework for semi-supervised object detection, which significantly enhances the performance of one-stage anchor-based detectors. The framework is based on a student-teacher mutual learning approach, as shown in Figure \ref{Fig.efficient_teacher_arch}, inspired by previous works \cite{xu2021end,chen2022label,chen2022dense,liu2021unbiased}. Our proposed Pseudo Label Assigner method divides pseudo labels into reliable and uncertain ones based on their scores, with reliable pseudo labels used for default supervised training, and uncertain ones used to guide the training of the student model with soft loss. The Epoch Adaptor method is used to speed up convergence by performing domain adaptation between labeled and unlabeled data, and calculating the threshold value of pseudo labels in each epoch. Throughout the training process, the teacher model employs the Exponential Moving Average (EMA) technique for updates.

\subsection{Dense Detector}
\begin{table}
  \resizebox{1.0\columnwidth}{!}{
  \begin{tabular}{l|ccccc}
    \toprule
    Method & Resolution & Mosaic & Param. & FLOPs & $AP_{50:95}(\%)$ \\
    \midrule
    Faster R-CNN \cite{ren2015faster} & [1333,800] && 39.8M & 202.31G & 40.3\\ 
    FCOS\cite{tian2019fcos}  & [1333,800] && 32.02M & 200.59G & 38.5\\
    YOLOv5 $w/o$ & [640,640] && 46.56M & 109.59G & 41.2\\
    \midrule
    YOLOv5\cite{jocher2022ultralytics} & [640,640] &$\checkmark$ & 46.56M & 109.59G & 49.0\\
    YOLOv7\cite{wang2022yolov7} & [640,640] &$\checkmark$ & 37.62M & 106.59G & 51.5\\
    \midrule
    RetinaNet\cite{lin2017focal} & [1333,800] && 37.74M & 239.32G & 39.5\\
    Dense Detector & [640,640] & $\checkmark$ & 42.13M & 169.61G & 44.86\\
    \bottomrule
  \end{tabular}}
  \caption{Comparison with Faster R-CNN, FCOS, YOLOv5, YOLOv7, RetinaNet and Dense Detector. The top section shows results for object detectors without Mosaic augmentation, the middle section shows results with Mosaic augmentation during training. Dense Detector  achieves comparable results to RetinaNet baseline, having lower FLOPs but greatly improved $AP_{50:95}$. Both Faster R-CNN, FCOS, RetinaNet and Dense Detector uses ResNet-50-FPN as backbone. $AP_{50:95}$ is reported on COCO val dataset.}
  \label{tab:detector_compare}
\end{table}
 YOLOv5\cite{jocher2022ultralytics} is a widely-used one-stage anchor-based detector in industry due to its friendly-deployed support and fast training speed. In order to investigate semi-supervised experiments on YOLOv5, a comprehensive analysis of the improvements made by YOLOv5 detector compared to other state-of-the-art detectors, such as RetinaNet, is required.  Results in Table~\ref{tab:detector_compare} demonstrate that YOLOv5 $w/o$ outperforms RetinaNet in terms of performance and computation. Furthermore, with dense image inputs after Mosaic augmentation, the $AP_{50:95}$ of YOLOv5 can be boosted from 41.2 to 49.0. YOLOv7 further improves the $AP_{50:95}$ to 51.5 with the help of dense flow of information and gradients on the basis of dense inputs. Based on the above comparison, a hypothesis can be derived that increasing the density of inputs can effectively enhance the performance of one-stage anchor-based detectors. To validate this hypothesis, a novel detector named Dense Detector was constructed to quantitatively evaluate the techniques employed in YOLOv5.

\begin{figure}[h]
    \centering 
    \begin{subfigure}[b]{0.25\textwidth}
        \centering
        \label{Fig.detector_compare.1}
        \includegraphics[width=35mm,height=75mm]{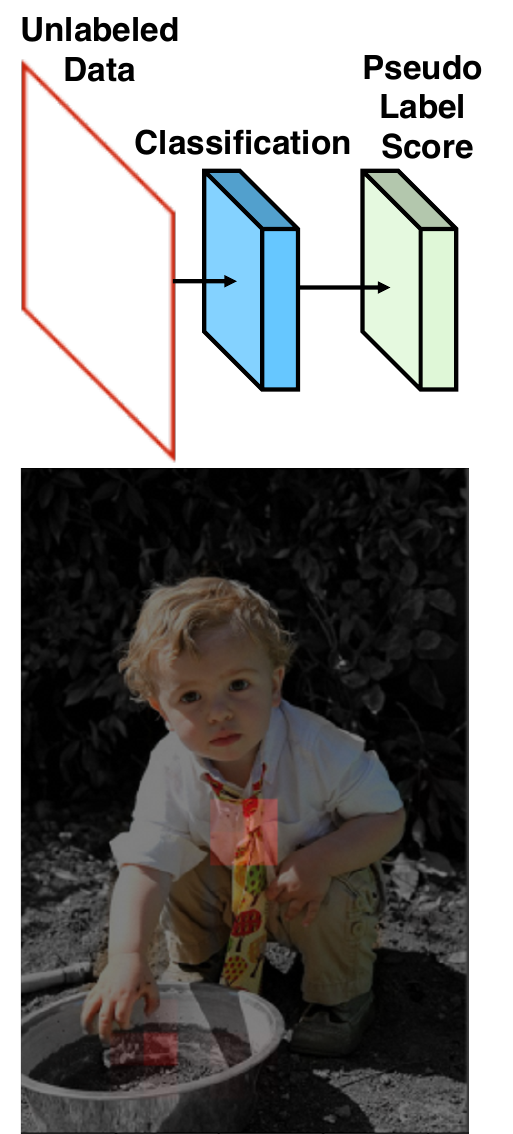}
        \caption{RetinaNet}
    \end{subfigure}
    \begin{subfigure}[b]{0.2\textwidth}
        \centering
        \label{Fig.detector_compare.2}
        \includegraphics[width=38mm,height=78mm]{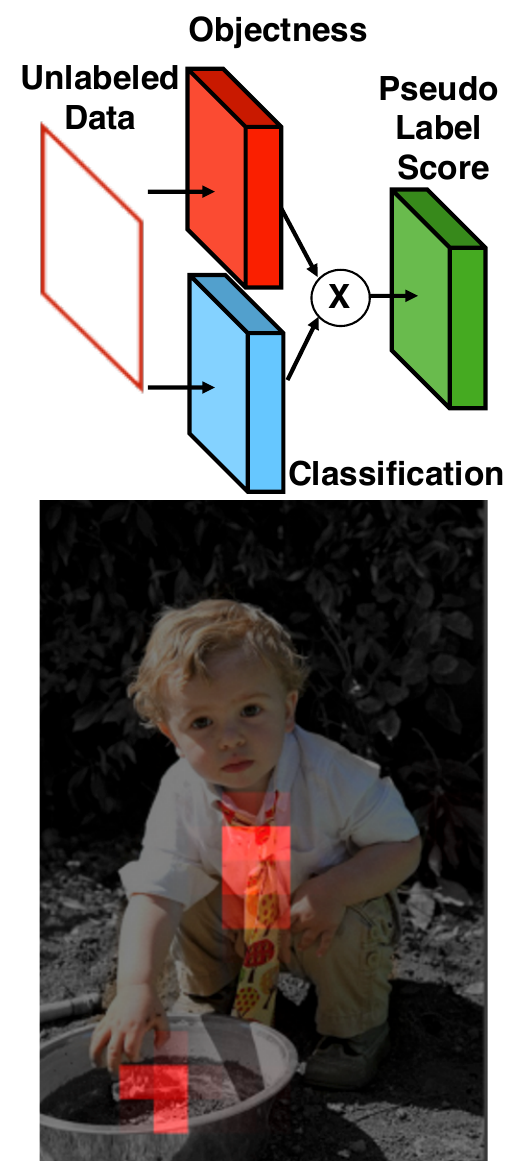}
        \caption{Dense Detector}
    \end{subfigure}
\caption{Comparison of pseudo label score heatmaps from RetinaNet and Dense Detector. Darker color indicates higher score. (a) RetinaNet produces sparse response due to the calculation of classification scores from pseudo labels generated from unlabeled data of 1333 $\times$ 800 input resolution. (b) Dense Detector, with input resolution of 640 $\times$ 640, uses a weighted pseudo label score based on objectness and classification scores, resulting in a more robust and dense response..}
\label{Fig.detector_compare}
\end{figure}

Dense Detector is modified from RetinaNet with ResNet-50-FPN backbone while changing the number of FPN output from 5 to 3, eliminating the weight sharing between detection headers and reducing the input resolution from 1333 to 640 for both training and inference. Additionally, Dense Detector has three output branches:a classification score , a bounding-box offset and an objectness score. Compared to RetinaNet, Dense Detetor achieved a 5.36\% $AP_{50:95}$ boost and 30\% faster inference, as reported in Table~\ref{tab:detector_compare}. Specifically, Dense Detector obtains objectness score by calculating the Complete Intersection over Union(CIoU)\cite{zheng2021enhancing} between the predicted and GT boxes. The Objectness score indicates the location quality of the predicted boxes and serves as an additional source of information to improve the detection performance. As illustrated in Figure \ref{Fig.efficient_teacher_arch}, the pseudo labels in SSOD are the predicted boxes of unlabeled data, the objectness scores of which indicate the location quality of pseudo labels. Thus, compared to RetinaNet with only a classification branch, Dense Detector with an extra objectness branch can indicate the location quality of pseudo labels during SSOD training as shown in Figure\ref{Fig.detector_compare}.


To verify the performance of Dense Detector in SSOD, 
we apply the classic SSOD method(Unbiased Teacher\cite{liu2021unbiased}) to the Dense Detector, which contains labeled and unlabeled data, teacher and student model, and a pseudo label filter to select pseudo labels. Furthermore, both labeled and unlabeled data branches adopt loss definition in Equation \ref{eq:supervised}.
However, in contrast to Unbiased Teacher on Faster R-CNN in Table~\ref{tab:coco standard}, the $AP_{50:95}$ improvement of Unbiased Teacher drops from 7.64 to 4.3 on Dense Detector. This motivated us to develop the following Pseudo Label Assigner that plays a key role in pseudo label assignment.

\subsection{Pseudo Label Assigner}
\begin{figure}[h]
    \centering 
    \includegraphics[width=85mm,height=65mm]{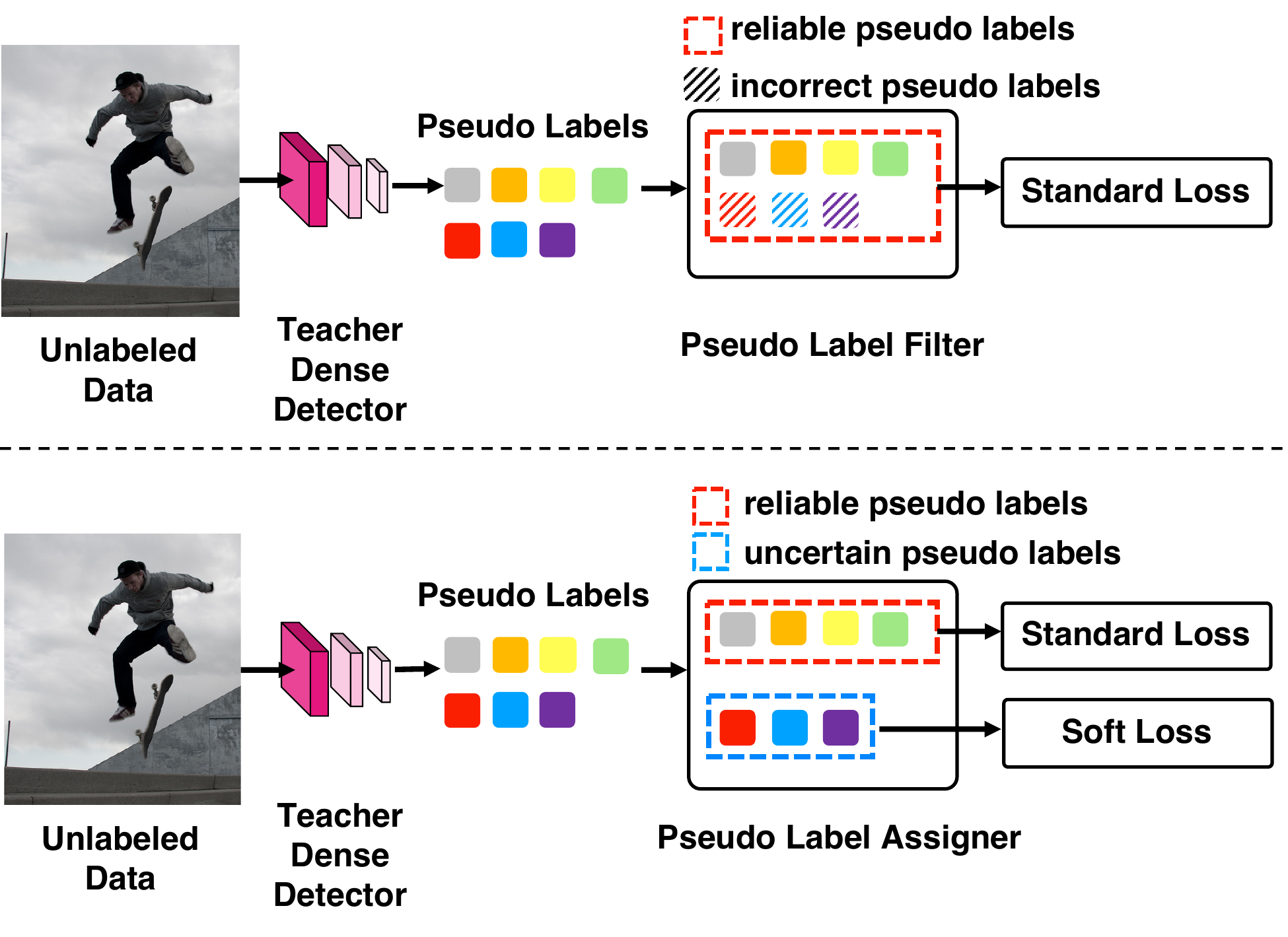}
\caption{Comparison of the impact of different pseudo label select strategy. In Pseudo Label Filter, a widely-used method in SSOD\cite{liu2021unbiased,xu2021end}, Setting the threshold too low can result in the generation of incorrect pseudo labels, while a threshold that is too high  may exclude reliable pseudo labels. This can lead to suboptimal assignments and adversely affect the training of the network. To address this issue, we propose the Pseudo Label Assigner method, which categorizes pseudo labels into reliable and uncertain categories based on high and low thresholds, respectively. The uncertain pseudo labels are assigned soft labels as targets for $L_u^{obj}$ to improve the quality of pseudo labels in SSOD.}
\label{Fig.pseudo_label_assigner}
\end{figure}

The core problem in SSOD is how to assign pseudo labels, as sub-optimal assignments can lead to inconsistent pseudo labels and deteriorating performance of the mutual learning mechanism. Pseudo Label Filter is a simple implementation for assigning labels, which filters out pseudo labels below a set threshold. Pseudo labels with scores below the threshold are labeled as background, while those with scores above the threshold are labeled as reliable pseudo labels. However, this method can result in sub-optimal assignments, as shown in Figure \ref{Fig.pseudo_label_assigner}: in the top case, Pseudo Label Filter is a fast method for filtering out pseudo labels with scores below the set threshold. However, during the entire process of SSOD training, the scores of pseudo labels continue to increase, which can lead to the Pseudo Label Filter treating incorrect pseudo labels as reliable ones and including them in training, resulting in the phenomenon of failure to converge in SSOD training.

Proposed in this work, Pseudo Label Assigner (PLA) provides a more refined assignment of the pseudo labels generated by Dense Detector. In PLA, pseudo labels obtained after Non-Maximum Suppression (NMS) are separated into two categories: reliable and uncertain pseudo labels. The high and low threshold ${\tau_{1},\tau_{2}}$ of the pseudo label score is used to determine two types of pseudo labels. Pseudo labels with scores between ${\tau_{1},\tau_{2}}$ are considered uncertain, and ignoring the loss of these labels directly results in improved performance on Dense Detector, as shown in Table \ref{tab:pseudo label assigner}. In addition to solving the sub-optimal problem caused by Pseudo Label Filter, PLA includes an unsupervised loss that efficiently leverages uncertain pseudo labels. The loss of Dense Detector in  SSOD is defined as a pair of single labeled image and single unlabeled image:
\begin{equation}
  L = L_s + \lambda L_u
\label{eq:total}
\end{equation}
\noindent
where $L_s$ represents the loss function computed on a labeled image, while $L_u$ represents the loss function computed on an unlabeled image,  $\lambda$ is used to balance the supervised loss and the semi-supervised loss, which is set to 3.0 in this paper. The $L_s$ follows the standard loss function in \cite{jocher2022ultralytics}:
\begin{equation}
\begin{aligned}
L_s & = \sum_{h,w}(CE(X_{(h,w)}^{cls}, Y_{(h,w)}^{cls}) + CIoU(X_{(h,w)}^{reg}, Y_{(h,w)}^{reg}) \\ & + CE(X_{(h,w)}^{obj}, Y_{h,w}^{obj}))
\end{aligned}
\label{eq:supervised}
\end{equation}
where CE indicates cross-entropy loss function, $X_{(h,w)}$ is the output of student model, and $Y_{(h,w)}$ means the sampled results generated by the label assigner of Dense Detector . The $L_u$ is defined as follows: 
\begin{equation}
\begin{aligned}
L_u = L^{cls}_u + L^{reg}_u + L^{obj}_u
\label{eq:L_u}
\end{aligned}
\end{equation}

\begin{equation}
\begin{aligned}
L^{cls}_u &= \sum_{h,w}(\mathbbm{1}_{\{p_{(h,w)}>=\tau_{2}\}}CE(X_{(h,w)}^{cls}, \hat{Y}_{(h,w)}^{cls}))
\label{eq:cls_u}
\end{aligned}
\end{equation}

\begin{small}
\begin{equation}
\begin{aligned}
L^{reg}_u = \sum_{h,w}(\mathbbm{1}_{\{p_{(h,w)}>=\tau_{2} \text{ or } \hat{obj}_{(h,w)} > 0.99 \}}CIoU(X_{(h,w)}^{reg}, \hat{Y}_{(h,w)}^{reg}))
\label{eq:reg_u}
\end{aligned}
\end{equation}
\end{small}

\begin{equation}
\begin{aligned}
L^{obj}_u &= \sum_{h,w}(\mathbbm{1}_{\{p_{(h,w)}<=\tau_{1}\}}CE(X_{(h,w)}^{obj}, \textbf{0})
\\& + \mathbbm{1}_{\{p_{(h,w)}>=\tau_{2}\}}CE(X_{(h,w)}^{obj}, \hat{Y}_{(h,w)}^{obj})) \\ & + \mathbbm{1}_{\{\tau_{1}< p_{(h,w)} <\tau_{2}\}}CE(X_{(h,w)}^{obj}, \hat{obj}_{(h,w)}) )
\label{eq:obj_u}
\end{aligned}
\end{equation}

\noindent
where $\hat{Y}^{cls}_{(h,w)}$, $\hat{Y}^{reg}_{(h,w)}$, $\hat{Y}^{obj}_{(h,w)}$ is the classification score, regression, objectness score of sampled results from PLA at location $(h,w)$ on feature map separately. $\hat{obj}_{(h,w)}$ is the objectness score of pseudo label at  $(h,w)$. $p_{(h,w)}$ is the score of pseudo label at  $(h,w)$. $\mathbbm{1}_{\{\cdot\}}$ is the indicator function, which outputs $1$ if condition $\{\cdot\}$ is satisfied and $0$ otherwise. 

  
\begin{figure*}[h]
    \centering 
    \begin{subfigure}[b]{0.3\textwidth}
        \centering
        \label{Fig.sub.1}
        \includegraphics[width=50mm,height=20mm]{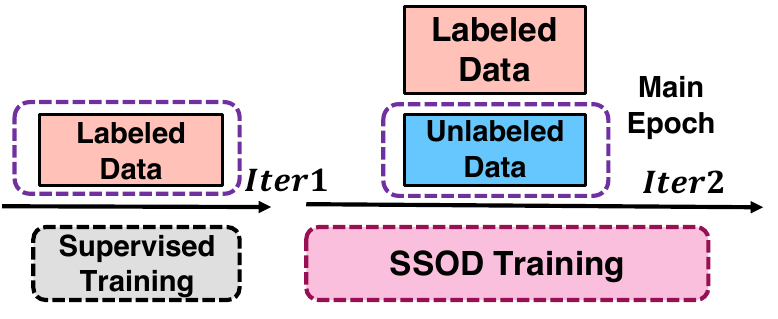}

        \caption{Alternating Training}
    \end{subfigure}
    \begin{subfigure}[b]{0.3\textwidth}
        \centering
        \label{Fig.sub.2}
        \includegraphics[width=45mm,height=20mm]{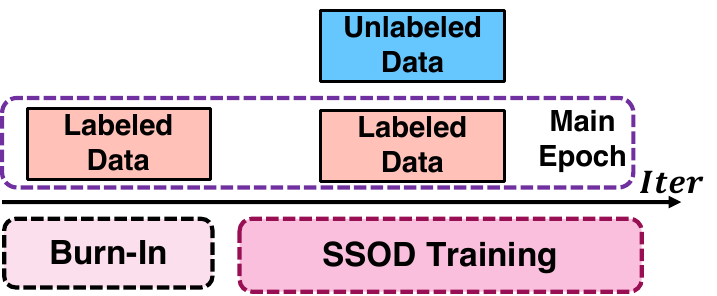}
        \caption{Joint Training with Burn-In}
    \end{subfigure}
    \begin{subfigure}[b]{0.3\textwidth}
    \centering
    \label{Fig.sub.3}
    \includegraphics[width=50mm,height=25mm]{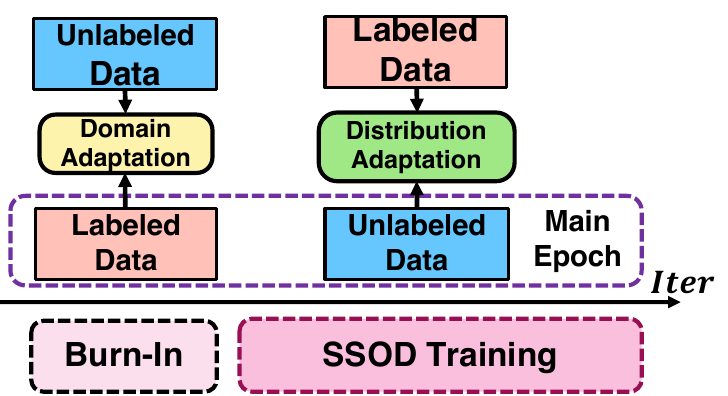}
    \caption{Joint Training with Epoch Adaptor}
    \end{subfigure}
\caption{Main epoch denotes a full training period that remains uninterrupted and without any reloading of new weights during its execution. Training strategies for Efficient Teacher: (a) supervised training on labeled data followed by SSOD training on unlabeled data; (b) supervised training on labeled data with additional SSOD training on unlabeled data; (c) end-to-end training on both labeled and unlabeled data with Epoch Adaptor incorporating Domain and Distribution Adaptation for improved convergence and feature distribution.}
\label{fig:epoch_adaptor_compare}
\end{figure*}

The difference between PLA and other pseudo labels selection strategies\cite{liu2022unbiased, chen2022dense} lies in that we designed a soft loss to handle uncertain pseudo labels separately. PLA distinguishes between two types of uncertain pseudo labels: those with high classification scores and those with high objectness scores. For the first type, only the objectness loss term $L_u^{obj}$ is computed. The targets of the cross-entropy $\hat{Y}{(h,w)}^{obj}$ are replaced with the soft label $\hat{obj}_{h,w}$, which indicates that these pseudo labels are not classified as either background or positive samples. For the second type, PLA calculates the regression loss term $L_u^{reg}$ when the objectness score is greater than 0.99. These pseudo labels have good regression results but insufficient classification scores to determine their label category. PLA aims to convert more uncertain pseudo labels into true positives using $L_u^{reg}$. This is important because during SSOD training on COCO, more than 70\% of uncertain pseudo labels are false positives due to inaccurate prediction boxes. Thus, PLA suppresses the inconsistency of pseudo labels through a soft label learning mechanism without affecting the loss of reliable pseudo labels. For more information, please refer to the Appendix.



\subsection{Epoch Adaptor}
While addressing the issue of pseudo label inconsistency in SSOD, the PLA still faces the challenge of achieving stability and high efficiency during training. To overcome this challenge, we introduce the Epoch Adaptor method, which leverages both domain adaptation and distribution adaptation techniques to enable rapid and stable SSOD training. Our approach aims to narrow the distribution gap between labeled and unlabeled data, while also dynamically estimating the threshold value for pseudo labels at each epoch.

As shown in the Figure~\ref{fig:epoch_adaptor_compare}, compared to alternating and original joint training scheme, EA enables the neural network to receive both labeled and unlabeled data during the Burn-In phase, employing domain adaptation techniques with a classifier to perplex the detector's capacity to discriminate between the two types of data. This effectively mitigates the overfitting effect that is commonly observed in the original approach, which relies solely on labeled data during the Burn-In phase. The domain adaptation loss function as follow: 
\begin{equation}
L_{da} = -\sum_{h,w}\Big[D\log p_{(h,w)} + (1-D)\log(1- p_{(h,w)})\Big].
\end{equation}
\noindent
where $p_{(h,w)}$ is the output of the domain classifier. $D$ = 0 for labeled data and $D$ = 1 for unlabeled data.We use the gradient reverse layer (GRL)~\cite{ganin2015unsupervised}, whereas the ordinary gradient descent is applied for training the domain classifier and the sign of the gradient is reversed when passing through the GRL layer to optimize the base network. In Burn-In phase, the supervised loss in one image can be reformulated as follows:
\begin{small}
\begin{equation}
\begin{aligned}
    L_s & = \sum_{h,w}(CE(X_{(h,w)}^{cls}, Y_{(h,w)}^{cls}) + CIoU(X_{(h,w)}^{reg}, Y_{(h,w)}^{reg}) \\ & + CE(X_{(h,w)}^{obj} + Y_{h,w}^{obj})) + \lambda L_{da}
\end{aligned}
\end{equation}
\end{small}
\noindent
where $\lambda$ is the hyper-parameter to control
the contribution of domain adaptation, which is 0.1 in our experiments. The expression capability of the model is enhanced by allowing the detector to see the unlabeled data in Burn-In. 

\begin{table*}[ht]
  \centering 
  \setlength{\tabcolsep}{1mm}{
   \resizebox{1.0\linewidth}{!}{

  \begin{tabular}{c|c|ccccc}
    \toprule
    \multicolumn{2}{c|} {Method} &  \%1 & \%2 & \%5 & \%10 & FLOPs\\
    \midrule
    \multirow{8}{*}{Two-stage anchor-based} & Supervised  & 9.05 & 12.70 & 18.47 & 23.86 & 202.31G \\
    & STAC\cite{sohn2020simple} & 13.97 $\pm$ 0.35\small{(\textcolor{blue}{$+4.92$})} & 18.25 $\pm$ 0.25 \small{(\textcolor{blue}{$+5.91$})} & 24.38 $\pm$ 0.12  \small{(\textcolor{blue}{$+5.91$})} & 28.64 $\pm$ 0.21 \small{(\textcolor{blue}{$+4.78$})} & 202.31G \\
    & Instant Teaching\cite{zhou2021instant} & 18.05 $\pm$ 0.15 \small{(\textcolor{blue}{$+9.00$})} & 22.45 $\pm$ 0.15 \small{(\textcolor{blue}{$+9.75$})} & 26.75 $\pm$ 0.05 \small{(\textcolor{blue}{$+8.28$})} & 30.40 $\pm$ 0.05 \small{(\textcolor{blue}{$+6.54$})} & 202.31G \\
    & Humber teacher\cite{tang2021humble} & 16.96 $\pm$ 0.38 \small{(\textcolor{blue}{$+7.91$})} & 21.72 $\pm$ 0.24 \small{(\textcolor{blue}{$+9.02$})} & 27.70 $\pm$ 0.15 \small{(\textcolor{blue}{$+9.23$})} & 31.61 $\pm$ 0.28 \small{(\textcolor{blue}{$+7.75$})} & 202.31G\\
    & Unbiased Teacher\cite{liu2021unbiased} & 20.75 $\pm$ 0.12  \small{(\textcolor{blue}{$+11.70$})} & 24.30 $\pm$ 0.07 \small{(\textcolor{blue}{$+9.80$})} & 28.27 $\pm$ 0.11 \small{(\textcolor{blue}{$+9.80$})} & 31.50 $\pm$ 0.10 \small{(\textcolor{blue}{$+7.64$})} & 204.13G \\
    & Soft Teacher\cite{xu2021end} & 20.46 $\pm$ 0.39 \small{(\textcolor{blue}{$+11.41$})} & - & 30.74 $\pm$ 0.08 \small{(\textcolor{blue}{$+12.27$})} & 34.04 $\pm$ 0.14  \small{(\textcolor{blue}{$+10.18$})} & 202.31G\\
    & LabelMatch\cite{chen2022label}& \textbf{25.81} $\pm$ 0.28 \small{(\textcolor{blue}{$+16.76$})} & - & 32.70 $\pm$ 0.18 \small{(\textcolor{blue}{$+14.23$})} & 35.49 $\pm$ 0.17 \small{(\textcolor{blue}{$+11.63$})} & 202.31G\\
    & PseCo\cite{li2022pseco} & 22.43 $\pm$ 0.36 \small{(\textcolor{blue}{$+13.38$})} & 27.77 $\pm$ 0.18 \small{(\textcolor{blue}{$+15.07$})} & 32.50 $\pm$ 0.08 \small{(\textcolor{blue}{$+14.03$})} & 36.06 $\pm$ 0.24 \small{(\textcolor{blue}{$+12.20$})} & 202.31G \\
    \midrule
    \multirow{4}{*}{One-stage anchor-free} & Supervised & 9.53 & 11.71 & 18.74 & 23.70 & 200.59G \\
    & Unbiased Teacher v2\cite{liu2022unbiased} & 22.71 $\pm$ 0.42 \small{(\textcolor{blue}{$+13.18$})} & 26.03 $\pm$ 0.12 \small{(\textcolor{blue}{$+14.32$})} & 30.08 $\pm$ 0.04 \small{(\textcolor{blue}{$+11.34$})} & 32.61 $\pm$ 0.03 \small{(\textcolor{blue}{$+8.91$})} & 200.59G\\
    & DSL\cite{chen2022dense} & 22.03 $\pm$ 0.28 \small{(\textcolor{blue}{$+12.50$})} & 25.19 $\pm$ 0.37 \small{(\textcolor{blue}{$+13.48$})} & 30.87 $\pm$ 0.24 \small{(\textcolor{blue}{$+12.13$})} & 36.22 $\pm$ 0.18 \small{(\textcolor{blue}{$+12.52$})} & 200.59G\\
    & Dense Teacher\cite{zhou2022dense} & 22.38 $\pm$ 0.31 \small{(\textcolor{blue}{$+12.85$})} & 27.20 $\pm$ 0.20 \small{(\textcolor{blue}{$+15.49$})} & 33.01 $\pm$ 0.21 \small{(\textcolor{blue}{$+14.27$})} & 37.13 $\pm$ 0.12 \small{(\textcolor{blue}{$+13.43$})} & 200.59G\\
    \midrule
     \multirow{5}{*}{One-stage anchor-based} & Supervised & 10.29 & 13.12 & 19.28 & 24.04 & 169.61G\\
    & Unbiased Teacher$\ast$\cite{liu2021unbiased} & 18.81 $\pm$ 0.28 \small{(\textcolor{blue}{$+9.07$})} & 22.72 $\pm$ 0.21 \small{(\textcolor{blue}{$+9.60$})} & 28.35 $\pm$ 0.12 \small{(\textcolor{blue}{$+8.15$})} & 30.34 $\pm$ 0.09 \small{(\textcolor{blue}{$+6.30$})} & 169.61G\\   
    & Ours  & 21.51 $\pm$ 0.21 \small{(\textcolor{blue}{$+11.22$})} & 27.15 $\pm$ 0.13 \small{(\textcolor{blue}{$+14.03$})} & 31.1 $\pm$ 0.08 \small{(\textcolor{blue}{$+11.82$})} & 34.09 $\pm$ 0.11 \small{(\textcolor{blue}{$+10.05$})} & 169.61G\\
    & Ours $\dagger$  & 23.76 $\pm$ 0.13 \small{(\textcolor{blue}{$+12.47$})} & \textbf{28.70} $\pm$ 0.14 \small{(\textcolor{blue}{$+15.58$})} & \textbf{34.11} $\pm$ 0.09 \small{(\textcolor{blue}{$+14.83$})}  & \textbf{37.90} $\pm$ 0.04 \small{(\textcolor{blue}{$+13.86$})} & 109.59G\\

    \bottomrule
  \end{tabular}
  }}
  \caption{Experimental results on COCO-standard ($AP_{50:95}$), $\ast$ means re-implemented
results on Dense Detector, $\dagger$ means Efficient Teacher with YOLOv5l\cite{jocher2022ultralytics}. All the results are the average of 5 folds.}
  \label{tab:coco standard}
\end{table*}
Moreover, current approaches require the calculation of the $\tau_1$ and $\tau_2$ thresholds of PLA for generating pseudo-labels on unlabeled data during SSOD training. Among these approaches, the most effective method\cite{chen2022label} entails leveraging the prior information of labeled data label distribution to compute the aforementioned thresholds. However, this method is not directly applicable to detectors such as Dense Detector, as we have demonstrated that Mosaic data augmentation plays an integral role in these detectors. Furthermore, the use of Mosaic data augmentation disrupts the label distribution ratio. To address this issue, we implement a distribution adaptation method based on the re-distribution method in LabelMatch \cite{chen2022label}. In distribution adaptation, the $\tau_1$ and $\tau_2$ thresholds at the $k$-th epoch are determined as follows:
\begin{small}
\begin{equation}
\begin{aligned}
    \tau_1^k = P_c^{k}[n_c^k\cdot \frac{N_u}{N_l}]
\end{aligned}
\end{equation}
\end{small}
\begin{small}
\begin{equation}
\begin{aligned}
    \tau_2^k = P_c^{k}[\alpha\%\cdot n_c^k\cdot \frac{N_u}{N_l}],
\end{aligned}
\end{equation}
\end{small}
\noindent
The reliable ratio $\alpha$ is set to 60 for all experiments, and $P_c^{k}$ represents the list of pseudo label scores of the $c$-th class at the $k$-th epoch. Meanwhile, $N_l$ and $N_u$ denote the number of labeled and unlabeled data, and $n_c^k$ represents the number of $c$-th class ground truth annotations that are counted by EA at the $k$-th epoch. By dynamically calculating the appropriate thresholds at each epoch, EA enables joint training to be more adaptable to dynamic data distributions. 

The integration of domain adaptation and distribution adaptation in EA effectively mitigates overfitting of neural networks to labeled data. Moreover, EA dynamically estimates appropriate thresholds for pseudo-labels at each epoch, achieving fast and efficient SSOD training. The experimental results demonstrating these effects are presented in Section 4.

\section{Experiments}
\subsection{Experimental Setup}
\textbf{Datasets.} We validate our method on MS-COCO\cite{lin2014microsoft} and VOC\cite{everingham2012pascal} benchmarks: (1) COCO-standard: 1\%, 2\%, 5\%, 10\% of the images are sampled on COCO as labeled data, and all the remaining data are used as unlabeled data. (2) COCO-additional: train2017 dataset is set as the labeled dataset and COCO2017-unlabeled is as the unlabeled dataset. (3) VOC: VOC07 trainval data is as the labeled dataset and VOC12 trainval is used as the unlabeled dataset. We adopt the mean average precision $AP_{50:90}$ as the evaluation metric.

\textbf{Network.}To verify that our proposed method is scalable, we used three Dense Detector architectures:The first one uses ResNet-50-FPN in Dense Detector. The second one replaces the original backbone with CSPNet and the Neck with PAN, which is similar with YOLOv5.

\textbf{Implementation Details.} We use 8 NVIDIA-V100 GPUs with 16G memory per GPU. We randomly sample 32 images from labeled data and 32 images from unlabeled data with ratio 1:1 in each iteration. For training configurations, the learning rate is 0.01 all the time, the $\tau_1$ and $\tau_2$ are calculated by EA. We used both weak and strong data augmentation. Mosaic is used in weak data augmentation. In the strong data augmentation, Mosaic, left-right flip, large scale jittering, graying, Gaussian blur, cutout, and color space conversion are selected. The max epoch is 300. Smoothing hyper-parameter in EMA is $0.999$. 

\subsection{Results}

\textbf{COCO-standard.} In Table ~\ref{tab:coco standard}, we validate our proposed method on COCO-standard and the performance of Efficient Teacher is better than Unbiased Teacher on Dense Detector. To ensure a fair comparison, we disabled the EMA of the supervised component during training of the supervised Dense Detector. This decision was made because none of both FasterRCNN and FCOS used EMA during training. Through this approach, we were able to conduct a precise evaluation of the effect of SSOD training on the detector's performance, which is quantified by the final gain. Our results indicate that the Efficient Teacher approach achieves a gain comparable to the state-of-the-art SSOD approach among Two-stage anchor-based and one-stage anchor-free SSOD methods. Furthermore, when we replaced the backbone of the Dense Detector with the standard YOLOv5l and trained it using Efficient Teacher, we observed a superior final detection performance with reduced computational overhead.

\textbf{COCO-additional.} Results in Table~\ref{tab:coco_additional} show our proposed method on COCO-additional, the gain effect of Efficient Teacher shows 1.45 increase on $AP_{50:95}$. The experimental results demonstrate that the performance of a YOLOv5l model, even when it has been trained to saturation, can be enhanced using Efficient Teacher. This improvement can be attributed to the inclusion of unlabeled data and pseudo labels, which mitigate overfitting on the labeled data and enable the model to learn a more generalized representation.
\begin{table}[h]
  \centering 
  \setlength{\tabcolsep}{1mm}{
  \begin{tabular}{ll}
    \toprule
    Method & $AP_{50:95}$\\
    \midrule
    Supervised $\dagger$ & 49.0 \\
    Ours $\dagger$ & \textbf{50.45}(\textbf{+1.45})\\
    \bottomrule
  \end{tabular}}
  \caption{Experimental results on COCO-additional.}
  \label{tab:coco_additional}
\end{table}

\textbf{PASCAL-VOC.} Table~\ref{tab:voc} shows the results of experiments conducted on VOC are convincing. Our method achieves 58.30 on $AP_{50:95}$. Moreover, since all other detectors were trained with an ImageNet pre-trained backbone, while ours was trained from scratch, we also report results using an ImageNet pre-trained backbone to initialize the Efficient Teacher. The Efficient Teacher with pre-trained backbone ultimately achieves superior SSOD training performance compared to its predecessor detector while utilizing only half the computational resources.
\begin{table}[h]
  \centering 
  \setlength{\tabcolsep}{1mm}{
  \begin{tabular}{lccc}
    \toprule
    Method & $AP_{50:95}$ & $AP_{50}$ & FLOPs\\
    \midrule
    STAC\cite{sohn2020simple} & 44.64 & 77.45 & 202.31G \\
    Instant Teacher\cite{zhou2021instant} & 50.00 & 79.20 & 202.31G\\
    Unbiased Teacher\cite{liu2021unbiased} & 48.69 & 77.37 & 204.13G\\
    Dense Teacher\cite{zhou2022dense} & 55.87 & 79.89 & 200.59G\\
    DSL\cite{chen2022dense} & 56.80 & 80.70 & 200.59G\\   
    Unbiased Teacher v2\cite{liu2022unbiased} & 56.87 & 81.29 & 200.59G\\
    LabelMatch\cite{chen2022label} & 55.11 & 85.48 &202.31G\\
    Ours $\dagger$ & 58.30 & 81.60 & 109.59G\\
    Ours $\ddagger$ & \textbf{60.56} & \textbf{86.54} & 109.59G\\
    \bottomrule
  \end{tabular}}
  \caption{Experimental results on PASCAL-VOC. The $\ddagger$ indicates using a ImageNet pre-trained backbone to initialize the Efficient Teacher }
  \label{tab:voc}
\end{table}

\subsection{Ablation Studies}
In ablation studies, we conducted experiments using the 10\% COCO-standard dataset(one of 5 folds). We set the backbone as the standard YOLOv5 because we have already analyzed the effective design of this detector in our previous experiments targeting the Dense Detector. Now, we will focus on verifying the specific effects of our proposed Efficient Teacher on the widely used YOLOv5l.


\textbf{Effect of Pseudo Label Assigner.} The impact of the proposed Pseudo Label Assigner is presented in Table~\ref{tab:pseudo label assigner}. We observe that applying the Unbiased Teacher method to the Dense detector with a threshold of 0.3 for pseudo label generation only leads to a modest $AP_{50:95}$ improvement of 1.65, which is considerably lower than the 7.6 $AP_{50:95}$ gain achieved by the Unbiased Teacher\cite{liu2021unbiased} applied to the Faster R-CNN. When neglecting the uncertain pseudo labels, the $AP_{50:95}$ further increases to 35.2. However, by utilizing the Pseudo Label Assigner to handle the uncertain pseudo labels, we obtain a significant improvement of 7.45 in $AP_{50:95}$, resulting in a final performance of 37.90, which is comparable to that of the Unbiased Teacher applied to the Faster R-CNN.
\begin{table}[h]
  \resizebox{1.0\columnwidth}{!}{
  \begin{tabular}{l|ll}
    \toprule
    Method & $AP_{50:95}$ & $AP_{50}$\\
    \midrule
    Supervised & 30.45 & 44.65 \\
    Unbiased Teacher\cite{liu2021unbiased} & 32.10 (+1.65) & 47.30 (+2.65)\\   
    Ignore uncertain pseudo label\cite{chen2022dense}  & 35.20 (+4.75)& 52.00 (+7.35)\\
    Pseudo Label Assigner & \textbf{37.90} \textbf{(+7.45)} & \textbf{54.19} \textbf{(+9.54)}\\
    \bottomrule
  \end{tabular}}
  \caption{Ablation study about different pseudo label assignment methods.}
  \label{tab:pseudo label assigner}
\end{table}

\textbf{Effect of Distribution Adaptation in EA.} We evaluate the impact of varying the threshold value $\tau_2$ in the Pseudo Label Assigner method on the COCO 10\% standard task. Table~\ref{tab:dynamic_thresh} demonstrates that an increase in $\tau_2$ leads to a reduction in $AP_{50:95}$, which suggests that fewer reliable pseudo labels and more uncertain ones are generated. This finding emphasizes the significance of maintaining an optimal balance between reliable and uncertain pseudo labels. Importantly, we observe that utilizing the distribution adaptation technique to dynamically compute the value of $\tau_2$ yields the best performance without requiring manual tuning. Our results suggest that this approach can lead to improved performance in SSOD training by striking an appropriate balance between reliable and uncertain pseudo labels, and by avoiding the negative impacts of manual tuning efforts.
\begin{table}[h]
  \centering 
  \setlength{\tabcolsep}{1mm}{
  \begin{tabular}{c|cc}
    \toprule
    $\tau_2$ & $AP_{50:95}$ & $AP_{50}$\\
    \midrule
    0.4 & 37.20 & 54.08 \\
    0.5 & 37.20 & 54.10 \\   
    0.6 & 36.90 & 53.77 \\
    0.7 & 35.10 & 51.60 \\
    \midrule
    EA & \textbf{37.90} & \textbf{54.80} \\
    \bottomrule
  \end{tabular}}
  \caption{Ablation studies on threshold value $\tau_2$ , EA indicates $\tau_2$ is calculated by Epoch Adaptor.}
  \label{tab:dynamic_thresh}
\end{table}

\textbf{Effect of the Domain Adaptation in EA.} Table~\ref{tab:domain adaptation} evidence that the utilization of domain adaptation leads to the improved convergence of SSOD training. This improvement can be attributed to domain adaptation effectively reducing the distributional divergence between labeled and unlabeled data, thus enabling more precise generation of pseudo-labels on the unlabeled data.
\begin{table}[h]
  \centering 
  \setlength{\tabcolsep}{1mm}{
  \begin{tabular}{l|ll}
    \toprule
    Method & $AP_{50:95}$ & $AP_{50}$\\
    \midrule
    w/o domain adaptation & 37.25 & 54.16\\
    domain adaptation & \textbf{37.90} & \textbf{54.80}\\
    \bottomrule
  \end{tabular}}
  \caption{Ablation studies on domain adaptation in EA.}
  \label{tab:domain adaptation}
\end{table}

\begin{figure}[h]
    \centering 
    \includegraphics[width=1\linewidth]{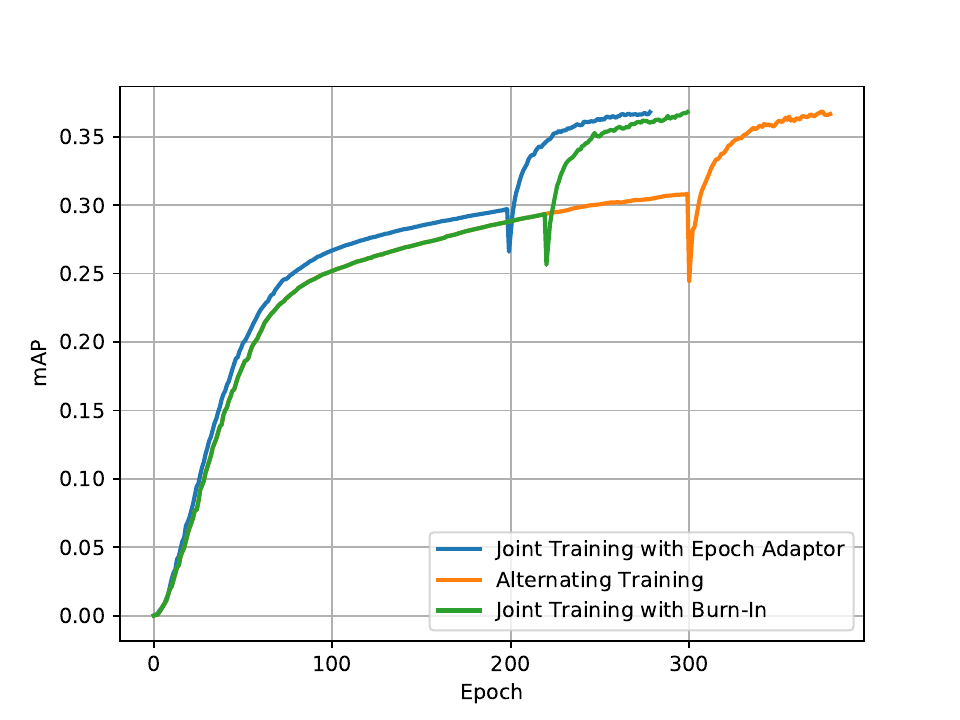}
    \caption{Performance ($AP_{50:95}$) comparisons of Epoch Adaptor, Alternating Training and Joint Training with Burn-In methods on COCO standard 10\%.}
    \label{fig:epoch_adaptor}
\end{figure}

\textbf{Ultimate impact of EA.} We demonstrated the accelerated training effect achieved by combining Domain Adaptation and Distribution Adaptation. (Figure \ref{fig:epoch_adaptor}). Our results demonstrate that joint training with Epoch Adaptor leads to superior performance with fewer iterations compared to fully supervised and alternating training. This highlights the potential of Epoch Adaptor as a more efficient and effective approach for training SSOD models.
\section{Conclusion}
In this paper, we present Efficient Teacher, a method to bridge the gap between SSOD and one-stage anchor-based detectors, by building on the efficient dense input handling of Dense Detector. Our approach introduces the Pseudo Label Assigner to effectively utilize both reliable and uncertain pseudo labels, based on an analysis of their assignment in SSOD. In addition, we introduce Epoch Adaptor, a training scheme that maximizes the efficiency of training and utilization of both labeled and unlabeled data. Efficient Teacher has been shown to achieve good SSOD results on various datasets, and has demonstrated both efficient training and deployment speeds. 
{\small
\bibliographystyle{ieee_fullname}
\bibliography{egbib}
}
\appendix

\end{document}